\documentclass[sigconf]{acmart}
\usepackage[T1]{fontenc}    
\usepackage{hyperref}       
\usepackage{url}            
\usepackage{booktabs}       
\usepackage{rotating}  
\usepackage{multirow} 
\usepackage{amsfonts}       
\usepackage{nicefrac}       
\usepackage{microtype}      
\usepackage{amsthm}
\usepackage{amsmath}
\usepackage{amsfonts}
\usepackage{bbm}
\usepackage{natbib}
\usepackage{graphicx}
\usepackage{subfig}
\usepackage[]{algorithm2e}
\usepackage{arydshln}
\usepackage{textcomp}
\usepackage{framed}
\usepackage{color} 

\usepackage{rotating}
\usepackage{arydshln}
\usepackage{wrapfig}

\copyrightyear{2018}
\acmYear{2018}
\setcopyright{rightsretained}
\acmConference[AIES '18]{2018 AAAI/ACM Conference on AI, Ethics, and
Society}{February 2--3, 2018}{New Orleans, LA, USA}
\acmBooktitle{2018 AAAI/ACM Conference on AI, Ethics, and Society (AIES '18),
February 2--3, 2018, New Orleans, LA, USA}
\acmDOI{10.1145/3278721.3278725}
\acmISBN{978-1-4503-6012-8/18/02}

\begin{document}
\title{Distill-and-Compare: Auditing Black-Box Models \\ Using Transparent Model Distillation}

\author{Sarah Tan}
\authornote{Work done while interning at Microsoft Research.}
\affiliation{%
  \institution{Cornell University}
}
\email{ht395@cornell.edu}

\author{Rich Caruana}
\affiliation{%
  \institution{Microsoft Research}
}
\email{rcaruana@microsoft.com}

\author{Giles Hooker}
\affiliation{%
  \institution{Cornell University}
}
\email{gjh27@cornell.edu}

\author{Yin Lou}
\affiliation{%
  \institution{Ant Financial}
}
\email{yin.lou@antfin.com}

\renewcommand{\shortauthors}{Tan, Caruana, Hooker, Lou}
\renewcommand{\shorttitle}{Auditing Black-Box Models Using Model Distillation and Comparison}

\begin{abstract}
Black-box risk scoring models permeate our lives, yet are typically proprietary or opaque. 
We propose Distill-and-Compare, a model distillation and comparison approach to audit such models. To gain insight into black-box models, we treat them as teachers, training transparent student models to mimic the risk scores assigned by black-box models. We compare the student model trained with distillation to a second un-distilled transparent model trained on ground-truth outcomes, and use differences between the two models to gain insight into the black-box model. Our approach can be applied in a realistic setting, without probing the black-box model API. We demonstrate the approach on four public data sets: COMPAS, Stop-and-Frisk, Chicago Police, and Lending Club. We also propose a statistical test to determine if a data set is missing key features used to train the black-box model. Our test finds that the ProPublica data is likely missing key feature(s) used in COMPAS. 
\end{abstract}



\maketitle
 
\section{Introduction}
\label{sec:introduction}
Risk scoring models have a long history of usage in criminal justice, finance, hiring, and other critical domains \cite{corbett2017algorithmic,louzada2016classification}. They are designed to predict a future outcome, for example defaulting on a loan. Worryingly, risk scoring models are increasingly used for high-stakes decisions, yet are typically proprietary or opaque.

Several approaches have been proposed \cite{henelius2014peek,Feldman2015Certifying, Adler2016AuditingBM, Adebayo2016Iterative, datta2016algorithmic, Wachter2018explanations} to audit black-box risk scoring models:
remove, permute, or obscure a protected feature, then see how the the model's predictions change after retraining the model or probing the model API with the transformed data. However, with many risk scoring models being proprietary, 
commercial model creators often do not provide unrestricted access to model APIs, much less release the model form or training data. 

In this paper, we study a more realistic setting where we only have a data set labeled with the risk score as produced by the risk scoring model, the information on the ground-truth outcome, and some or all features; we are not able to probe the model API with new data. We call this data set the \textit{audit data}. We add two potential complications: the audit data may not be the original training data, and the audit data may not have all features used to train the risk scoring model. For example, ProPublica obtained data for their COMPAS study \cite{propublica_story} not from the company that created COMPAS, but through a public records request to Broward County, a US jurisdiction that used COMPAS in their criminal justice system \cite{propublica_method}, and ProPublica may not have had access to all the features Broward County used to get COMPAS scores. 

We propose a two-step approach to audit black-box risk scoring models, using audit data with both black-box risk scores and ground-truth outcomes. First, we use a Distill-and-Compare approach (Section \ref{sec:distillation}) to understand how features (in the audit data) affect the risk scores. Then, we use a statistical test (Section \ref{sec:correlation}) to determine if the black-box model used additional features we do not have access to (i.e. features not in the audit data). 

The contributions of this paper are: 1) We propose an approach using model distillation and comparison to audit black-box risk scoring models under realistic conditions.
2) We show the importance of calibrating risk scores to remove scale distortions that may have been introduced during the training of the black-box model.
3) We apply the approach to audit four risk scoring models.
4) We propose a statistical test to determine if the audit data is missing key features used to train the black-box model. 
5) An ancillary contribution of this paper is a new confidence interval estimate for iGAM \cite{lou2012intelligible, lou2013accurate, caruana2015intelligible}, a type of transparent model, to compare two models of this class. 

\section{Audit Approach}
\label{sec:approach}
Our goal is to gain insight into the input-output relationships of a black-box risk scoring model. We draw from model distillation and model comparison techniques to develop our approach.

\subsection{Distill and Compare}
\label{sec:distillation}
Model distillation was first introduced to transfer knowledge from a large, complex model (teacher) to a faster, simpler model (student) \cite{bucilua2006compression,hinton2014distilling,ba2014deep}. This was done by running unlabeled samples (either new unlabeled data or training data with labels discarded) through the teacher model to obtain the teacher's outputs, then training the student model to mimic the teacher's outputs. We draw parallels to our setting, taking the risk scoring model to be the teacher and the audit data to be unlabeled samples ran through the teacher (risk scoring model) to obtain the teacher's output (risk scores). We train the mimic model to minimize mean squared error between the teacher and student, i.e. 
\vspace{-0.2cm}
\begin{equation}
L(S, \hat{S}) =  \frac{1}{T} \sum_{t=1}^T \left(S(x^t) - \hat{S}(x^t)\right)^2
\end{equation}
 where $x^t$ is the t-$th$ sample in the audit data, $S(x^t)$ is the output of the teacher model (risk scores) for sample $x^t$, $\hat{S}(x^t)$ is the output of the mimic model for sample $x^t$, and $T$ is the number of samples. Throughout this paper, we will call the teacher model the \textit{black-box model} and the student model the \textit{mimic model}. 

Next, we leverage the ground-truth outcome information. We train \textit{our own risk scoring model} on the audit data to predict the ground-truth outcome, i.e.
\vspace{-0.2cm}
\begin{align}
L(O, \hat{O}) &= \frac{1}{T}\sum_{t=1}^T \Bigl\{O(x^t)\log\left(P(\hat{O}(x^t)=1)\right) + \nonumber \\ &\quad(1-O(x^t))\log\left(P(\hat{O}(x^t)=0)\right)\Bigr\}
\end{align}
 where $O(x^t) \in \{0,1\}$ is the ground-truth outcome for sample $x^t$ and $\hat{O}(x^t) \in \{0,1\}$ is the output of the model for sample $x^t$. Throughout this paper, we call this model the \textit{outcome model}. Note that the outcome model is not a mimic model. 
 
  \begin{figure}[htb]
   \centering
     \includegraphics[width=0.9\linewidth]{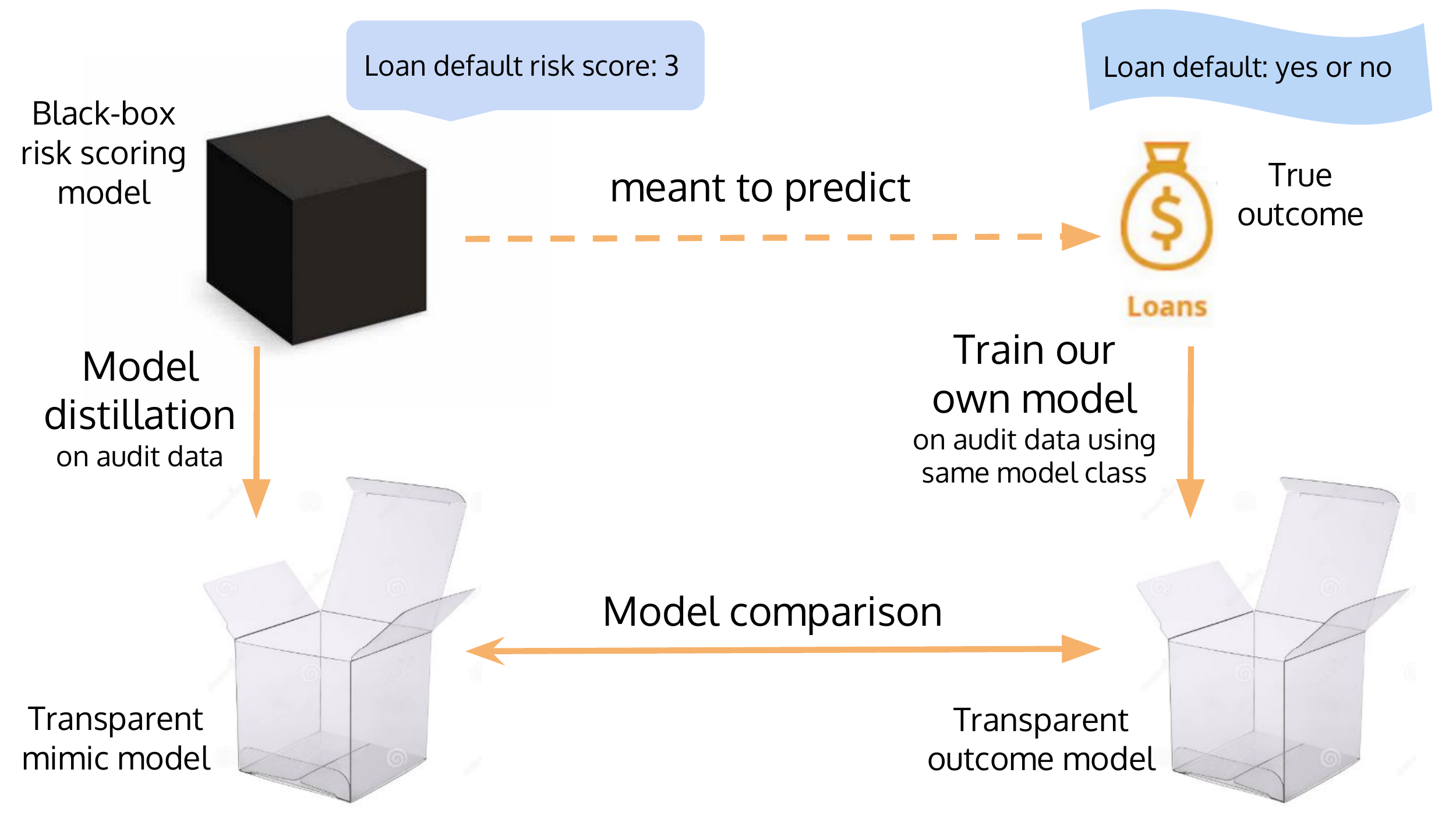}
     \caption{Auditing a loan risk scoring model by training transparent models on data labeled with the risk scores and with ground-truth outcomes for loan defaults.}
     \label{fig:distillation}
      \vspace{-0.35cm}
 \end{figure}
 
 It is critical that both the mimic model and outcome model are trained using the same model class that allows for interpretation and comparison.
 Not all model classes have the property that two models of that class can be compared. E.g. it is not obvious how to compare two decision trees, random forests or neural nets. We want a model class that is as rich and complex as possible so that the mimic model can be faithful to the black-box model and the outcome model can accurately predict ground-truth outcomes. But this model class should still be transparent \cite{doshivelez2017towards} so that we can understand how the input features affect the models' outputs---the goal of our audit. We focus on a particular transparent model class, iGAM (Section \ref{sec:modelclass}) in this paper; other choices are possible.   
 
The risk score and the ground-truth outcome are closely related---the ground-truth outcome is what the black-box model was meant to predict. If the black-box model is accurate \textit{and} generalizes to the audit data, it would predict the ground-truth outcomes in the audit data correctly; the converse is true if the black-box model is not accurate \textit{or} does not generalize to the audit data. 

Because both the mimic and outcome models are trained with the same model class on the same audit data using the same features, the more faithful the mimic model, and the more accurate the outcome model, the more likely it is that differences observed between the mimic and outcome models result from differences between the ground-truth outcomes and risk scores assigned by the black-box risk scoring model. Moreover, similarities between the mimic and outcome models (e.g. on COMPAS in Section \ref{sec:COMPAS}, the Number of Priors feature is modeled very similarly by the two models) increases confidence that the mimic model is a faithful representation of the black-box model, and that any differences observed on other features are meaningful. 

\vspace{0.1cm}
\noindent \textbf{Related work.} 
Adler et al. \cite{Adler2016AuditingBM} trained a model to predict outcomes and then a second model to predict the first's predictions. This is a different distillation setup from ours, as we use both risk scores and outcomes. Adebayo and Kagal \cite{Adebayo2016Iterative} also learned their own risk scoring models when the black-box model cannot be queried. Some papers also compare two models, but not risk scores and outcomes at the same time. Wang et al. \cite{wang2018direction} trained a model to predict outcomes and another to predict membership in a protected subgroup. Chouldechova and G’Sell \cite{Chouldechova2017Fairer} trained two different outcome models then identified subgroups where the two models differed. Other papers work on a single outcome or risk score model \cite{Zhang2017Predictivebias,tramer2017fairtest}.

\subsection{Testing for Missing Features}
\label{sec:correlation}
If the audit data is missing features used by the black-box model, the audit data alone may be insufficient to audit the black-box model. We propose a statistical test to test if there are important missing features based on the following observation:

\begin{quote}
\emph{If the black-box model used features that are missing from the audit data but are useful for predicting the ground-truth outcome, the error between the mimic model (learned on the audit data) and the risk score, $||\hat{S}-S||_E$, should be positively correlated with the error between the outcome model (learned on the audit data) and ground-truth outcome, $||\hat{O}-O||_E$.
}
\end{quote}

\begin{figure*}[htb]
\begin{minipage}{.48\textwidth}
\begin{turn}{90}\hspace{0.001cm} $\quad$ {\fontsize{3}{4}\selectfont Feature Contribution, $h_i(x_i)$}\end{turn}
\includegraphics[width=0.235\textwidth,height=0.8in]{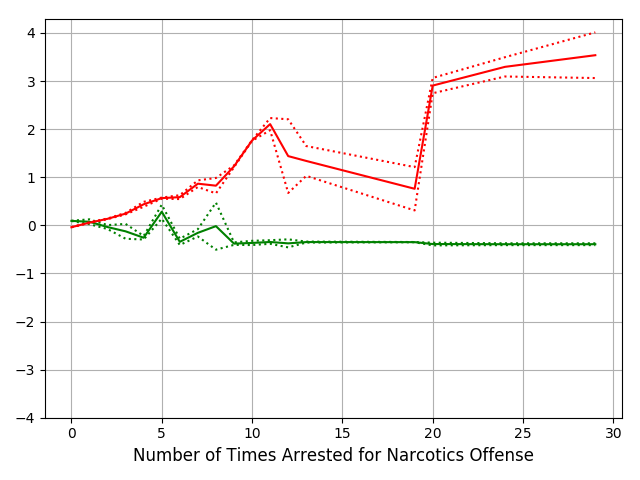}
\includegraphics[width=0.235\textwidth,height=0.8in]{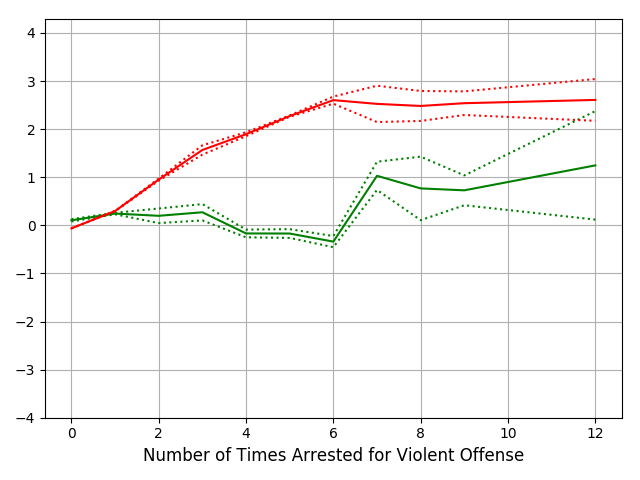}
\includegraphics[width=0.235\textwidth,height=0.8in]{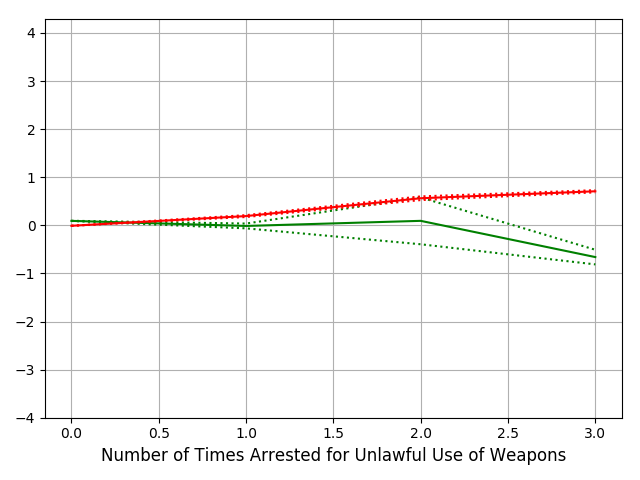}
\includegraphics[width=0.235\textwidth,height=0.8in]{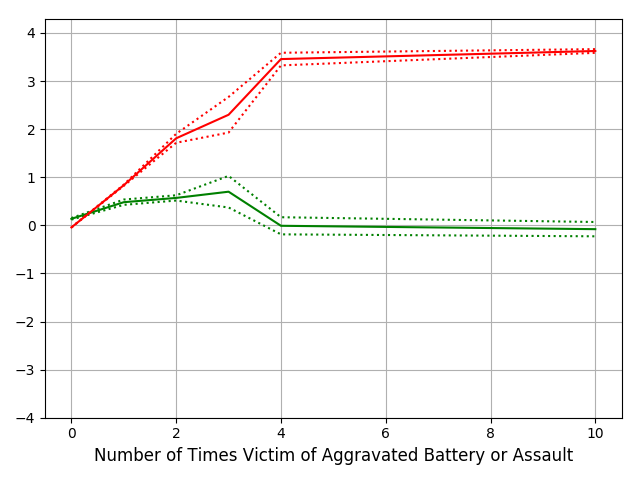}
\begin{turn}{90}\hspace{0.001cm} $\quad$ {\fontsize{3}{4}\selectfont Feature Contribution, $h_i(x_i)$}\end{turn}
\includegraphics[width=0.235\textwidth,height=0.8in]{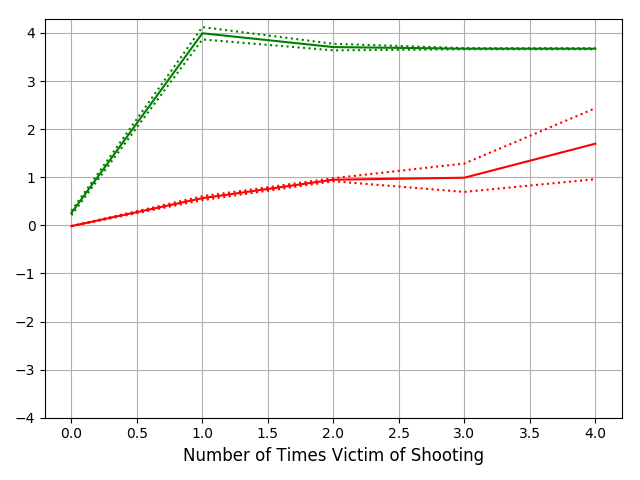}
\includegraphics[width=0.235\textwidth,height=0.8in]{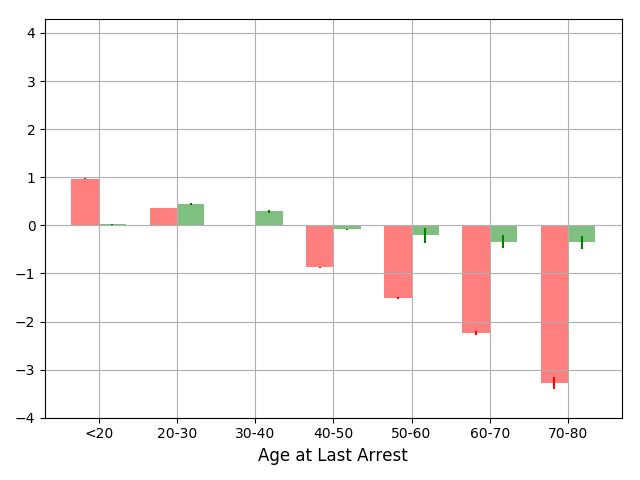}
\centering
\includegraphics[width=0.235\textwidth,height=0.8in]{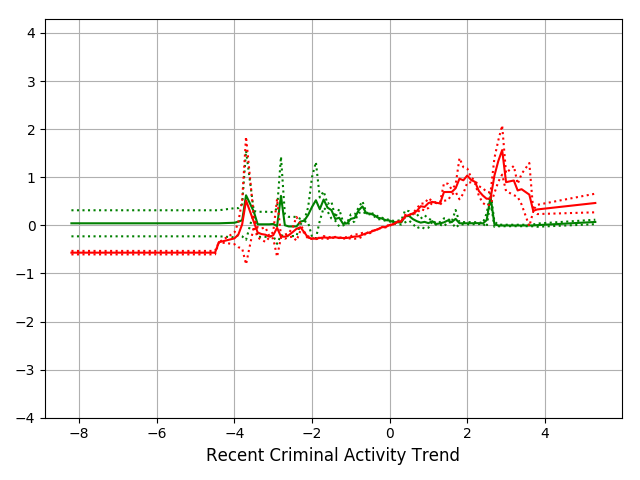}
\includegraphics[width=0.235\textwidth,height=0.8in]{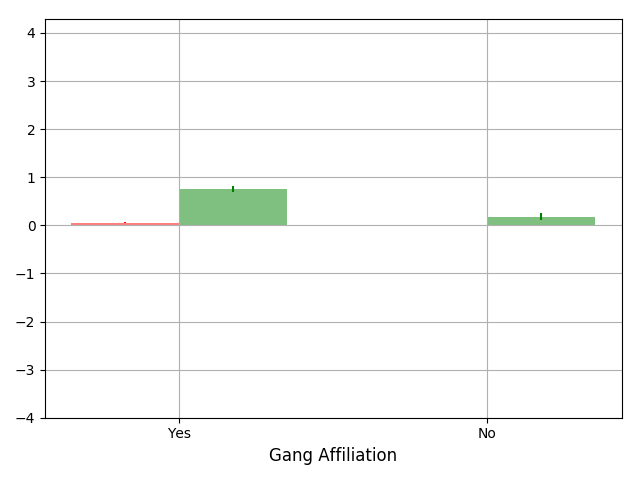}
\vspace*{-0.05in}
\caption{Eight features the Chicago Police says are used in their risk scoring model.  Best seen on screen.
}
\label{fig:police_feats}
\end{minipage}
\hspace*{0.1in}
\begin{minipage}{.48\textwidth}
\begin{turn}{90}\hspace{0.001cm} $\quad$ {\fontsize{3}{4}\selectfont Feature Contribution, $h_i(x_i)$}\end{turn} \includegraphics[width=0.235\textwidth,height=0.8in]{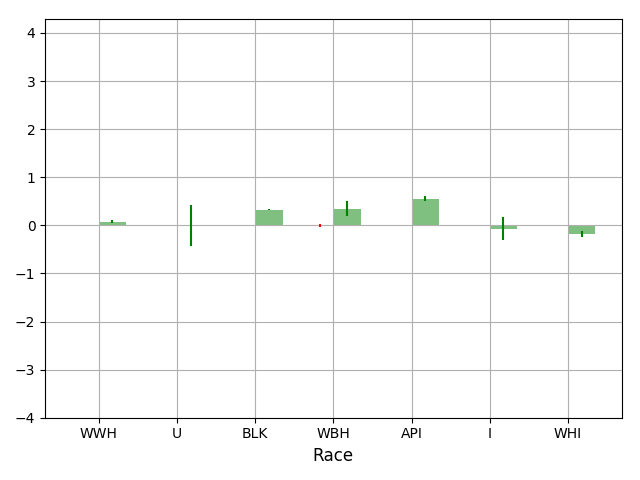}
\includegraphics[width=0.235\textwidth,height=0.8in]{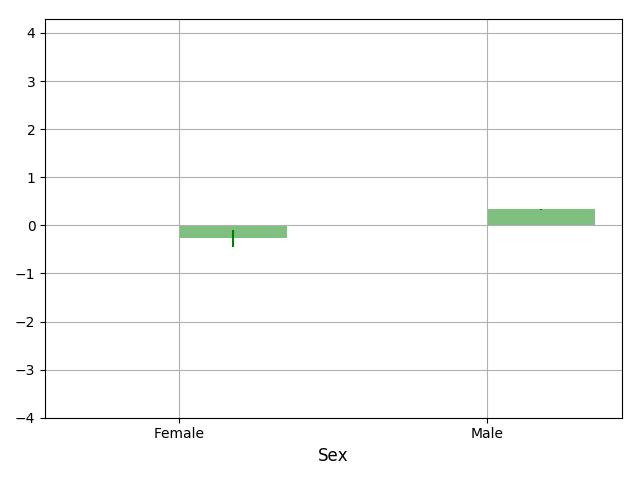}
\includegraphics[width=0.235\textwidth,height=0.8in]{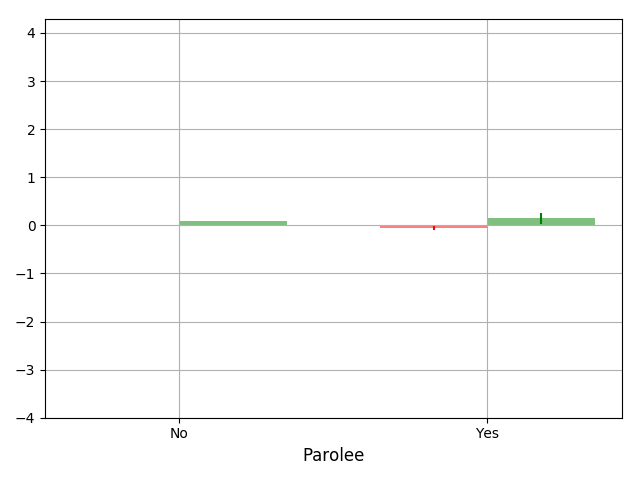}
\includegraphics[width=0.235\textwidth,height=0.8in]{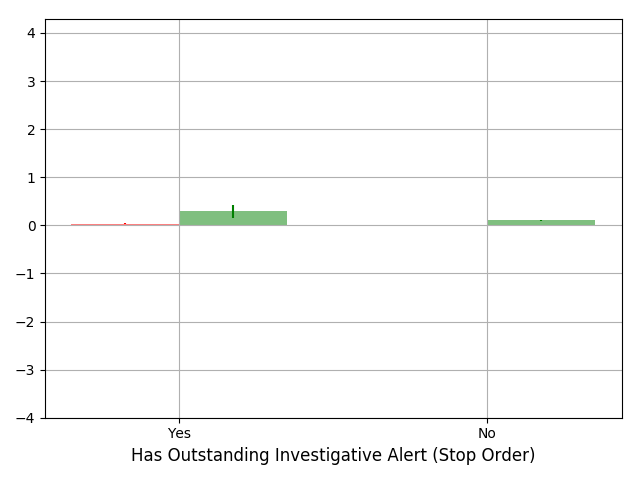}
\begin{turn}{90}\hspace{0.001cm} $\quad$ {\fontsize{3}{4}\selectfont Feature Contribution, $h_i(x_i)$}\end{turn}
\includegraphics[width=0.235\textwidth,height=0.8in]{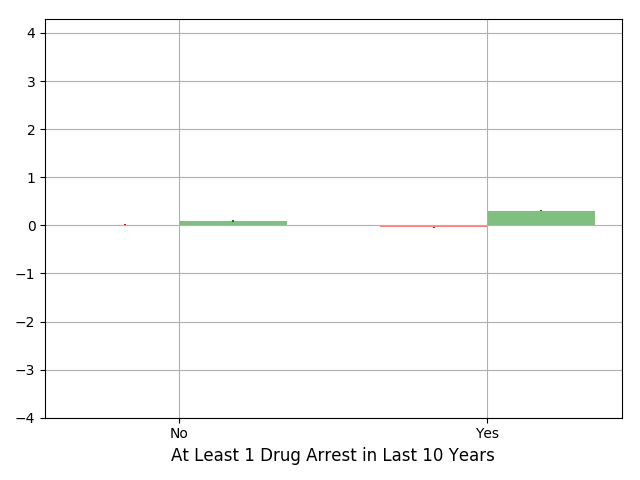}
\includegraphics[width=0.235\textwidth,height=0.8in]{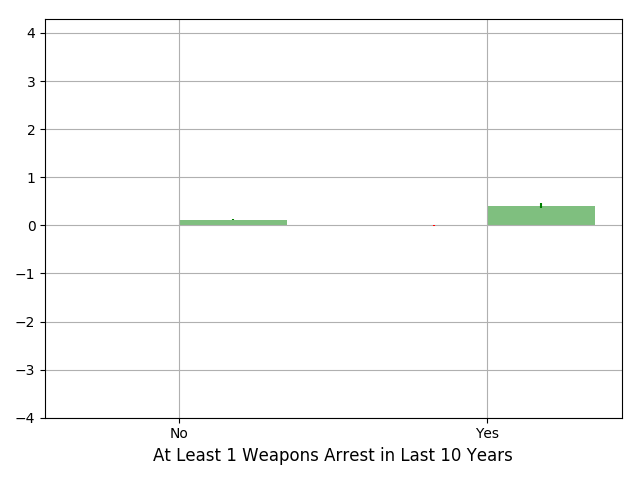}
\centering
\includegraphics[width=0.235\textwidth,height=0.8in]{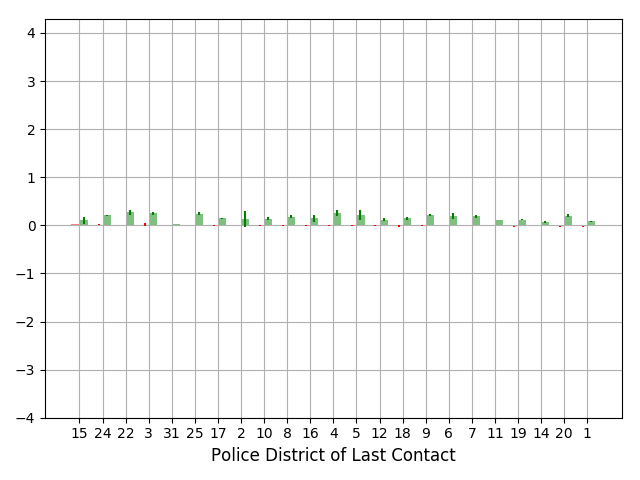}
\includegraphics[width=0.235\textwidth,height=0.8in]{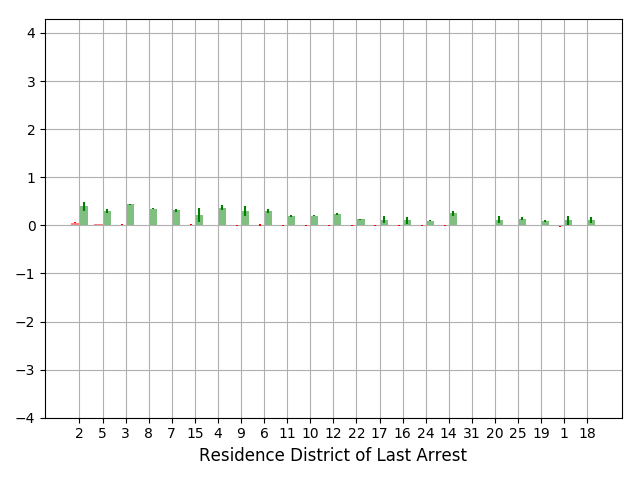}
\vspace*{-0.05in}
\caption{Eight features the Chicago Police says are \emph{not used} in their risk scoring model. Best seen on screen.
}
\label{fig:police_not_feats}
\end{minipage}
\end{figure*}

Since the test involves predictions from the mimic and outcome models, both models need to be trained (Sections \ref{sec:distillation}) prior to performing this test; this test merely checks if the results of our audit are significantly affected by missing features. We perform this test on all the risk scoring models we audit in this paper in Section \ref{sec:missing}. Independent of our audit approach, this test can standalone as a test for whether a data set is missing key features that were used to train a black-box model. Transparency of the mimic and outcome models is not a requirement for this test.  

\subsection{Comparing Mimic and Outcome Models}
\label{sec:compare}
In this section, we provide technical details on how to train the mimic and outcome models so that they are comparable, and detect statistically significant differences.

\subsubsection{Choice of model class} 
\label{sec:modelclass}
As noted in Section \ref{sec:distillation}, we train the mimic model and outcome model using the same transparent model class---in this paper, iGAM \cite{lou2012intelligible, lou2013accurate, caruana2015intelligible}. We point the reader to \cite{lou2012intelligible, lou2013accurate, caruana2015intelligible} to learn more about iGAMs and to \cite{tan2018transparent} for a distillation example where it was used as a student. Briefly, iGAM has the form
\begin{equation} \label{eq:ga2m}
g(y) = h_0 + \sum_i h_i(x_i) + \sum_{i \neq j} h_{ij}(x_i,x_j) \end{equation} where the contribution of any one feature $x_i$ or pair of features $x_i$ and $x_j$ to the prediction can be visualized in graphs such as Figure~\ref{fig:police_feats} that plot $x_i$ on the x-axis and $h_i(x_i)$ on the y-axis. For classification, $g$ is the logistic function. For regression, $g$ is the identity function. For classical GAMS \cite{hastie1990generalized}, feature contributions $h(\cdot)$ are fitted using splines; for iGAM, they are fitted using ensembles of short trees. Crucially, since iGAM is an additive model, two iGAM models can be compared by simply taking a difference of their feature contributions $h(\cdot)$, which we exploit in Section \ref{sec:differences} to detect differences between the mimic and outcome models.

\subsubsection{Calibrating model inputs}
\label{sec:calibration}
Calibration is the process of matching predicted and empirical probabilities \cite{degroot1983reliability, niculescu05predicting}. If a risk score is well-calibrated, the relationship between the risk score and empirical probabilities will be linear (e.g. COMPAS and Stop-and-Frisk in the top row of Figure \ref{fig:reliability} in the Appendix). While developing our approach, we discovered that not all risk scores exhibit the desired linear relationship with outcomes in the audit data. For example, the Chicago Police risk score (third column of Figure \ref{fig:reliability} in the Appendix) is rather flat for risk scores less than 350, then exhibits a sharp kink upwards.

One possible explanation for any nonlinear relationship is that the risk score was well-calibrated on its original training data, but the audit data has a different distribution (data shift) \cite{riley2016external}. Another explanation is post-processing by risk scoring model creators to reduce sensitivity in less important parts of the risk score scale and enhance separation in more important parts of the scale \cite{lingo2008discriminatory}. 

We make the reasonable assumption that risk scores should be monotonic and well-calibrated \cite{lingo2008discriminatory} and use this assumption to undo distortions or audit data shift. Specifically, we learn a nonlinear transformation of the risk score (the blue line in Figure \ref{fig:reliability} in the Appendix), similar to isotonic regression \cite{niculescu05predicting}, to make the risk scores and outcomes linearly related on a scale of choice. The mimic model is then trained on these transformed risk scores. This calibration step is necessary to compare the mimic and outcome models, as it makes the targets for the two models (risk scores and outcomes) linearly related on a scale that their outputs will later be compared on. We select this scale to be logit probabilities (since the outcome model's outputs are already on this scale), and perform this calibration step for Chicago Police and Lending Club but not COMPAS and Stop-and-Frisk, since the latter two already exhibit the desired linear relationships. 

\subsubsection{Detecting differences}
\label{sec:differences}
To not mistake random noise for real differences between the mimic and outcome models, we control potential sources of noise during the training process. To avoid data sample-specific effects, we train the mimic and outcome models on the same data sample. We then calculate the difference in feature $x_i$'s contribution to the models, 
$sh_i(x_i) - oh_i(x_i)$. If this number is positive, the mimic model assigns more risk than the outcome model for feature $x_i$; the converse is true if this number is negative. 

We construct a confidence interval for this difference to tell if it is statistically significant. One ancillary contribution of this paper is a new method to estimate confidence intervals for the iGAM model class, by employing a {\em bootstrap-of-little-bags} approach \cite{sexton2009standard} to estimate the variance of $h_i(x_i)$ and $sh_i(x_i)-oh_i(x_i)$. See Appendix \ref{sec:variance} for details. The resulting confidence intervals are the dotted lines in Figures \ref{fig:police_feats}, \ref{fig:police_not_feats}, and \ref{fig:recid_feats}.

\section{Results}
\label{sec:results}

\subsection{Validating the Audit Approach}
\label{sec:validating}
\begin{figure*}[htb]
\begin{center}
\begin{tabular}{ccccc}
\begin{turn}{90}\hspace{0.5cm} $\quad$ \scriptsize{Feature Contribution, $h_i(x_i)$}\end{turn}
\includegraphics[width=0.235\textwidth,height=1.6in]{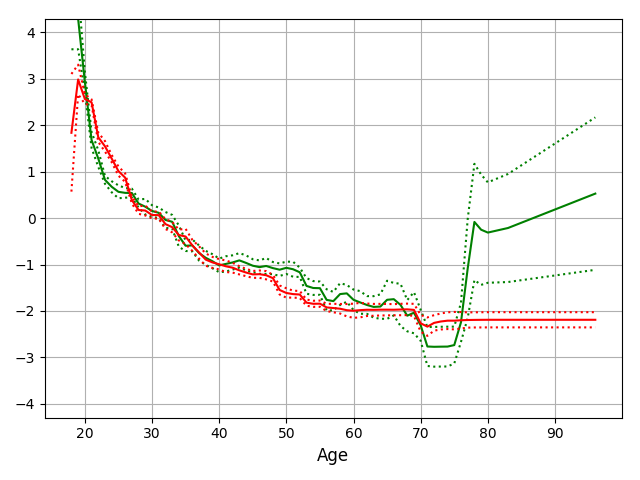} & \hspace*{-0.1in}
\includegraphics[width=0.235\textwidth,height=1.6in]{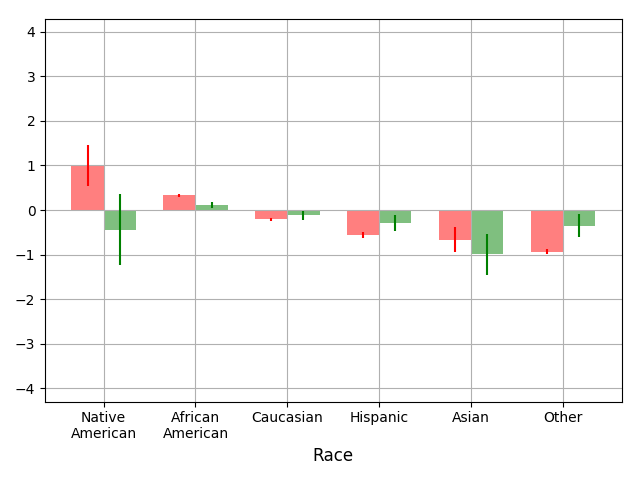} & \hspace*{-0.1in}
\includegraphics[width=0.235\textwidth,height=1.6in]{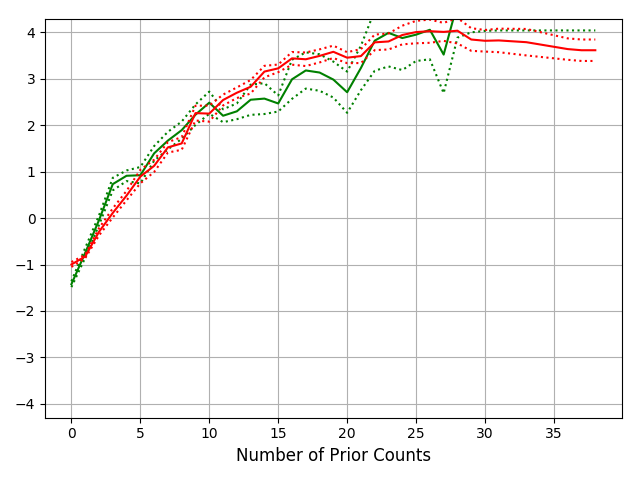} & \hspace*{-0.1in} 
\includegraphics[width=0.235\textwidth,height=1.6in]{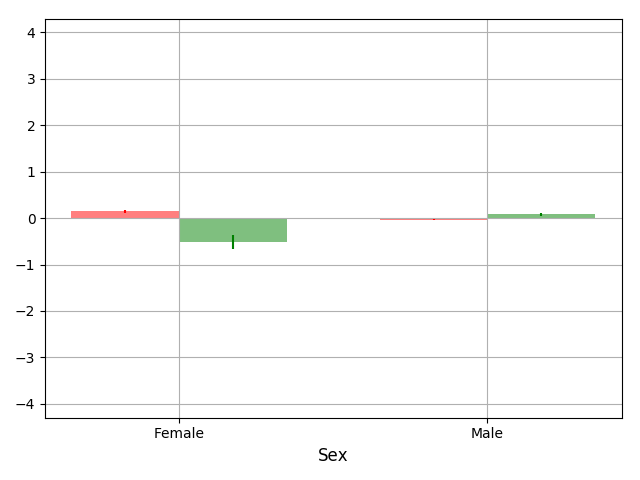} & \hspace*{-0.1in} 
\end{tabular}
\end{center}
\vspace*{-0.175in}
\caption{Feature contributions of four features to the COMPAS mimic model (in red) and outcome model (in green). 
}
\label{fig:recid_feats}
\vspace{-0.1cm}
\end{figure*}

In this section, we demonstrate the approach on risk scoring models where we have some information on how they were trained, and check that our approach can recover this information. 

\subsubsection{Stop-and-Frisk.}
\label{sec:stopfrisk}
Using the New York Police Department's Stop-and-Frisk data\footnote{http://www1.nyc.gov/site/nypd/stats/reports-analysis/stopfrisk.page}, Goel et al. \cite{goel2016precinct} proposed a simple risk scoring model for weapon possession: $S = 3 \times \mathbbm{1}_{PS} + 1 \times \mathbbm{1}_{AS} + 1 \times \mathbbm{1}_{Bulge}$,
where $S$ is the risk score, $PS$ denotes primary stop circumstance is presence of suspicious object, $AS$ denotes secondary stop circumstance is sight or sound of criminal activity, and $Bulge$ denotes bulge in clothing \cite{goel2016precinct}.
Since we know the model's functional form, we can verify that the mimic model correctly recovers its coefficients.
We apply the risk scoring model to label 2012 data (T=126,457, p=40) after following Goel et al.'s data pre-processing steps \cite{goel2016precinct}. 

\vspace{0.1cm}
\noindent \textbf{Result.} The mimic model recovers the coefficients of (3, 1, 1) for the three features used in the risk scoring model ($PS$, $AS$, $Bulge$) and 0 for the remaining features.

\subsubsection{Chicago Police ``Strategic Subject''.}
The Chicago Police Department released arrest data\footnote{\url{https://data.cityofchicago.org/Public-Safety/Strategic-Subject-List/4aki-r3np}} from 2012 to 2016 that was used to create a risk score for the probability of an individual being involved in a shooting incident as a victim or offender. 
This data set contains 16 features, but only 8 are used by the black-box model, which gives us an opportunity to test if our approach can accurately detect which features are and are not used by a black-box model. 

We trained a mimic model, intentionally including all 16 features. Figure~\ref{fig:police_feats} shows the feature contributions of the mimic model (in red) and outcome model (in green) for the 8 features the Chicago Police says were used by the black-box model; Figure~\ref{fig:police_not_feats} shows the 8 features the Chicago Police says were \textit{not used} in their model. 

\vspace{0.1cm}
\noindent \textbf{Result.} There is a striking difference between Figure~\ref{fig:police_feats} and Figure~\ref{fig:police_not_feats}: the mimic model (in red) assigns importance to the features in Figure \ref{fig:police_feats}, but does not assign any importance to the features in Figure \ref{fig:police_not_feats}. This agrees with Chicago Police's statement about which features were and were not used in the black-box model. We also note that the outcome model (in green) does assign importance to the unused features (Figure \ref{fig:police_not_feats}), suggesting that there is signal available in the 8 unused features that the Chicago Police risk scoring model could have used, but chose not to use. Race and sex are 2 of these 8 features, which the Chicago Police especially emphasize are not used. 

These experiments confirm that the mimic models can provide insights into the black-box models, and demonstrate the advantages of using the outcome information available in the audit data.

\subsection{Auditing COMPAS}
\label{sec:COMPAS}
COMPAS, a proprietary score developed to predict recidivism risk, has been the subject of scrutiny for racial bias \cite{propublica_story, kleinberg2016inherent, chouldechova2017fair, corbett2017algorithmic, blomberg2010validation, dieterich2016COMPAS}.
We do not know what model class, input features or data were used to train COMPAS. As described in Section \ref{sec:introduction}, the COMPAS audit data\footnote{\url{https://github.com/propublica/COMPAS-analysis}} was collected by ProPublica; it is likely different from the original COMPAS training data. 
Figure~\ref{fig:recid_feats} compares the COMPAS mimic model (in red) and outcome model (in green) for four features: Age, Race, Number of Priors, and Gender. The dotted lines are 95\% pointwise confidence intervals. We observe the following:

\vspace{0.1cm}
\noindent\textbf{COMPAS agrees with ground-truth outcomes regarding the number of priors.} In the 3rd plot in Figure \ref{fig:recid_feats}, the mimic model and outcome model agree on the impact of Number of Priors on risk; their confidence intervals overlap through most of its range.

\vspace{0.1cm}
\noindent\textbf{COMPAS disagrees with ground-truth outcomes for some age and race groups.} The 1st and 2nd plots in Figure \ref{fig:recid_feats} show the effect of Age and Race on the mimic and outcome models. The mimic model (red) and the outcome model (green) are very similar between ages 20 to 70, and their confidence intervals overlap. For ages greater than 70, there is evidence that the models disagree as the confidence intervals do not overlap. 

The mimic and outcome models are also different for ages 18 and 19: the mimic model predicts low risk for young individuals, but we see no evidence to support this in the outcome model, with risk appearing to be highest for young individuals.

The mimic model predicts that African Americans are even higher risk, and Caucasians lower risk, than the ground-truth outcomes suggest is warranted. Note that the ground-truth outcomes might themselves be biased due to historical discrimination against African Americans. 

\vspace{0.1cm}
\noindent\textbf{Gender has opposite effects on COMPAS compared to ground-truth outcome.} In the 4th plot in Figure \ref{fig:recid_feats}, we see a discrepancy between the mimic model and outcome model on Gender. The mimic model predicts higher risk than warranted by ground-truth outcomes for females, and conversely for males. 

\vspace{0.1cm}
\noindent \textbf{Explaining differences between mimic and outcome models.} Each observed difference between the two models could have several different explanations. We propose to leverage the observed differences to gain insight into the black-box model, by asking the question, ``what could be happening in the black-box model, that could explain the differences we are seeing''? We provide several examples below.

\begin{figure}[htb]
\begin{center}
\begin{tabular}{ccccc}
\begin{turn}{90}\hspace{0.5cm} $\quad$ \scriptsize{Feature Contribution, \scriptsize{$h_{ij}(x_i,x_j)$}}\end{turn}
\includegraphics[width=0.235\textwidth,height=1.6in, trim={0.05cm 0.3cm 0.5cm 1cm},clip]{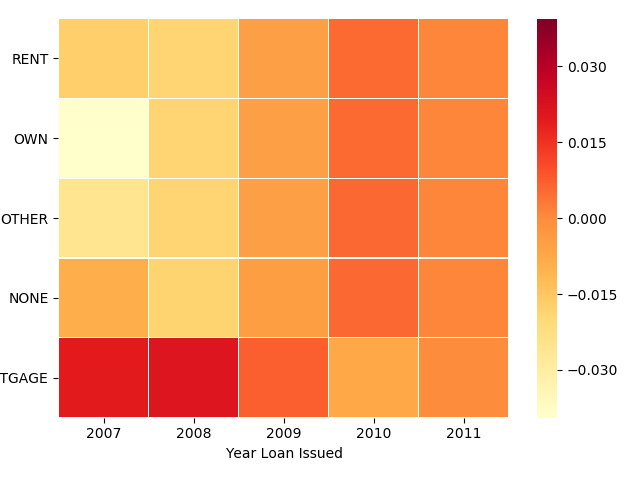} & \hspace*{-0.1in}
\includegraphics[width=0.235\textwidth,height=1.6in, trim={0.05cm 0.3cm 0.5cm 1cm},clip]{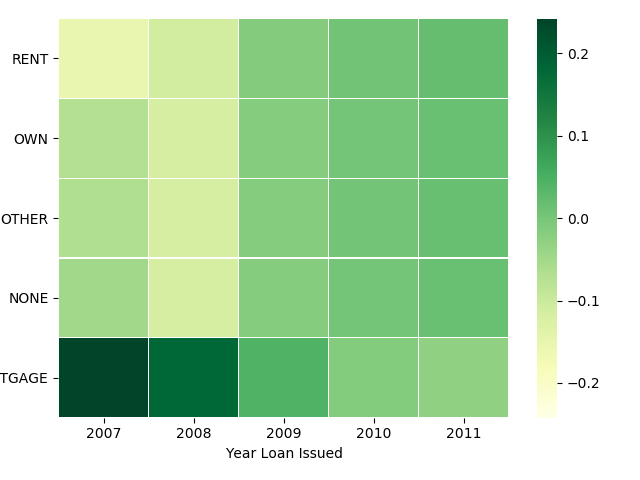} & \hspace*{-0.1in}
\\
\end{tabular}
\end{center}
\vspace*{-0.175in}
\caption{Interaction between loan issue year and home ownership in  Lending Club mimic model (in red) and outcome model (in green).
}
\label{fig:loan_feats}
\vspace{-0.35cm}
\end{figure}

\begin{enumerate}
    \item Some feature regions may be underrepresented in the black-box model's training data and/or the audit data. For example, in the COMPAS audit data (collected by ProPublica), only 3\% of samples are between 18 and 20 years old, only 0.5\% are older than 70 years old, and only 19\% are female, which makes learning accurate models in these regions harder. 
\item The black-box model may have a more simple or complex form than the outcome model. For example, we saw that the mimic model predicts low risk for young individuals, but there is no evidence to support this in the outcome model. We trained an iGAM model with interactions between pairs of features, and observed strong interactions between very young age and other variables such as Gender, Charge Degree, and Length of Stay. If COMPAS does not model interactions well, this may explain why COMPAS needs to predict low risk for very young individuals (because it cannot otherwise predict a reduced risk via interactions of age with other variables). 
\item The black-box model may be deliberately simple for some feature regions. For example, for ages greater than 70, the outcome model has much wider confidence intervals than the mimic model. The ground-truth outcomes are potentially high-variance in this region, yet the black-box model's scoring function may have been kept deliberately simple for extreme feature values like this. 
\item The black-box model may be using additional features missing from the audit data, that interact with the non-missing features. We provide a statistical test to determine the likelihood of this (Section \ref{sec:missing}). 
\end{enumerate}
and so on. We cannot tell (without further testing) the definitive reason that explains a particular difference between the mimic and outcome models. However, we suggest that this form of reasoning coupled with the use of transparent models surfaces differences that we did not \textit{a priori} know, but can then proceed to check and reason about, to gain insight into the black-box model.

\subsection{Auditing Lending Club}
Lending Club, an online peer-to-peer lending company, rates loans it finances and releases data\footnote{\url{https://www.lendingclub.com/info/download-data.action}} on these loans. We use a subset of five years (2007-2011) of loans that have matured, so that we have ground-truth on whether the loan defaulted. We do not know what model class or input features Lending Club used to train their risk scoring model. We believe the data sample we have is similar to the data they would have used to train their models. According to Lending Club, their models are refreshed periodically.

We use this Lending Club example to discuss an insight we gained into the black-box model from examining interactions revealed by our transparent models. Figure~\ref{fig:loan_feats} shows the interaction of loan issue year and home ownership in the Lending Club risk scoring mimic model (in red) and ground-truth outcome model (in green). Having a home mortgage in 2007-2008 increases the loan default risk more than having a home mortgage in 2009 and beyond. Recall that 2007-2008 is around the time of the subprime housing crisis. Note the difference in ranges between the two plots---the range goes up to 0.2 for the outcome model (in green) whereas the range is much lower for the mimic model (in red). This could indicate that the Lending Club risk scoring model is updated conservatively, with some lag time, instead of being rapidly updated as economic conditions and behavior change.

\subsection{Testing for Features Missing from \\ Audit Data}
\label{sec:missing}
As black-box models may use  additional features we do not have access to, we developed a test (Section \ref{sec:correlation}) to assess the impact missing features could have on our analysis. Table~\ref{table: missingfeatures} provides 95\% confidence intervals for three linear and nonlinear measures of correlation used in the test. If zero is in the confidence interval, the error of the mimic model (trained on the audit data) is not correlated to the error of the outcome model (also trained on the audit data). Then, it is unlikely that the audit data is missing key feature(s) that are a) predictive of outcomes (and hence will negatively affect the error of the outcome model if missing); and b) used in the black-box model (and hence will negatively affect the error of the mimic model if missing). 

\begin{table}[t!]
\centering
\caption{Statistical test for likelihood of audit data missing key features used by black-box model.}
\label{table: missingfeatures}
\vspace{-0.15cm}
\begin{tabular}{@{}llll@{}}
\toprule
\textbf{Risk Score}  & \textbf{Pearson} $\rho$  & \textbf{Spearman} $\rho$   & \textbf{Kendall} $\tau$\\
\toprule
COMPAS & [0.10, 0.13] & [0.10, 0.14] & [0.08, 0.10] \\
Lending Club & [0.00, 0.03] & [-0.01, 0.01] & [-0.01, 0.01] \\
Stop-and-Frisk & [0.00, 0.01] & [-0.03, 0.01] & [-0.02, 0.01] \\
Chicago Police & [0.00, 0.01] & [0.01, 0.03] & [0.01, 0.02] \\
\bottomrule      
\end{tabular}
\vspace{-0.3cm}
\end{table}

In Lending Club and Stop-and-Frisk we cannot distinguish these correlations from zero, suggesting that no key features are missing from the audit data. For Chicago Police, the confidence intervals contain 0 or are very close to 0 (lower limit 0.01), hence there is little evidence of the audit data missing key features. For COMPAS there is evidence of positive correlation, indicating that the ProPublica data may be missing key features used in the COMPAS model. This is supported by our findings in Section \ref{sec:fidelity} that no mimic models trained on the ProPublica data, however powerful (e.g. random forests), could mimic COMPAS well.  

\begin{table*}[t!]
\centering
\caption{Fidelity of mimic model and accuracy of outcome model. Lower RMSE is better, higher AUC is better.} 
\label{table: studenterror}
\vspace{-0.15cm}
\begin{tabular}{lllllll}
\toprule
& \textbf{Risk Score}  & \textbf{Metric}  & \textbf{Linear model}   & \textbf{iGAM}  & \textbf{iGAM w/ interactions} & \textbf{Random Forest} \\
\toprule
\multirow{4}{*}{\shortstack[c]{Fidelity\\ of mimic \\model}} & COMPAS
& RMSE (1-10)
& $2.11 \pm 0.057$
& $2.01 \pm 0.045$
& $\mathbf{2.00 \pm 0.047}$
& $2.02 \pm 0.053$ \\
& Lending Club &  RMSE (2-36)
& $3.27 \pm 0.037$
& $2.60 \pm 0.049$
& $2.52 \pm 0.051$
& $\mathbf{2.48 \pm 0.033}$\\
& Chicago Police &  RMSE (0-500)
& $17.4 \pm 0.102$
& $17.2 \pm 0.125$
& $16.5 \pm 0.130$
& $\mathbf{14.0 \pm 0.280}$ \\
& Stop-and-Frisk
& RMSE (0-5)
& $\mathbf{0.00\pm  2\times10^{-15}}$
& $\mathbf{0.00 \pm 1\times10^{-5}}$
& $\mathbf{0.00 \pm 2\times10^{-5}}$
& $0.01 \pm 2\times10^{-3}$\\
\midrule
\multirow{4}{*}{\shortstack[c]{Accuracy \\of outcome\\ model}} & COMPAS
&  AUC
& $0.73 \pm 0.029$
& $0.74 \pm 0.027$
& $\mathbf{0.75 \pm 0.029}$
& $0.73 \pm 0.026$\\
& Lending Club
&  AUC
& $0.69 \pm 0.006$
& $\mathbf{0.69 \pm 0.016}$
& $\mathbf{0.69 \pm 0.014}$
& $0.68 \pm 0.020$ \\
& Chicago Police
&  AUC
& $\mathbf{0.95 \pm 0.007}$
& $\mathbf{0.95 \pm 0.007}$
& $\mathbf{0.95 \pm 0.007}$
& $0.929 \pm 0.009$ \\
& Stop-and-Frisk
&  AUC
& $0.84 \pm 0.020$
& $0.85 \pm 0.020$
& $0.85 \pm 0.020$
& $\mathbf{0.87 \pm 0.024}$ \\
\bottomrule
\end{tabular}
\end{table*}

\subsection{Fidelity and Accuracy}
\label{sec:fidelity}
To quantitatively evaluate our audit approach, we report fidelity (how well the mimic model predicts the black-box model's risk scores, measured in RMSE) and accuracy (how well the outcome model predicts the ground-truth outcomes, measured in AUC) for all the risk scoring models we audit in Table~\ref{table: studenterror}. For comparison, we also train linear models (a simpler model class than iGAM) and random forests (more complex, but less interpretable). 

For COMPAS, all model classes (linear model, iGAM, random forest) have roughly equal fidelity and accuracy. Interestingly, none obtained RMSE lower than 2 on a 1-10 scale.  
Comparing outcome model AUCs across different model classes, iGAM's results are generally comparable to (or slightly better than) more complex random forests (Table \ref{table: studenterror}). For the risk score mimic models, random forests are competitive for Lending Club and Chicago Police. Linear mimic models are not far behind iGAMs for several risk scoring models (COMPAS, Chicago Police, Stop-and-Frisk), suggesting that the black-box model's functional form might be very simple. We know this to be true for Stop-and-Frisk from Section~\ref{sec:stopfrisk} where the model was a simple linear model. 

\vspace{0.1cm}

\noindent \textbf{COMPAS deep dive.} One possible reason why COMPAS is challenging to mimic may be that the ProPublica data set is missing key features. This agrees with the results of the statistical test in Section \ref{sec:missing}. Another possible reason is the small sample size (less than 7,000 samples). 

One advantage of using a model distillation approach to inspect black-box models is that the approach may be able to benefit from additional unlabeled data if the black-box teacher model can be queried to label the additional data \cite{bucilua2006compression}. 
We found an additional 3,000 individuals in the ProPublica data with COMPAS risk scores \textit{but no ground-truth outcomes}. Adding them to the training (not testing) data for the mimic model and retraining the mimic model, we find marginal improvement in the mimic model's fidelity (from RMSE 2.0 to 1.98). Doing the opposite---removing individuals from the training data in 1,000 increments---decreased the mimic model's fidelity only marginally (to RMSE 2.1, training on only 1,000 individuals). These analyses suggest that for COMPAS, missing key features is a more pressing issue than insufficient data.

\section{Discussion}

Sometimes we are interested in detecting bias on features intentionally excluded from the black-box model.  For example, a credit risk scoring model is probably not allowed to use race as an input.  Unfortunately, not using race does not prevent the model from learning to be biased.  Racial bias in a data set is likely to be in the outcomes --- the labels used for learning; not using race as an input \emph{feature} does not remove the bias from the \emph{labels}. 

If race were uncorrelated with all other features (and combinations of features) provided to the model, then removing race would prevent the model from learning to be racially biased because it would not have any input features on which to model this bias.  Unfortunately, in any large, real-world data set, there is massive correlation among the high-dimensional features, and a model trained to predict credit risk will learn to be biased from correlation of the \textit{excluded} race feature with other features that likely remain in the model (e.g., income or education). 

Hence, removing a protected feature like race or gender does not prevent a model from learning to be biased. Instead, removing protected features make it harder to detect how the model is biased, or correct the bias, because the bias is now spread in a complex way among all the correlated features throughout the model instead of being localized to the protected features.  The main benefit of excluding protected features like race or gender from the inputs of a machine learning model is that it allows the group deploying the model to claim (incorrectly) that their model is not biased because it did not use these features.

When training a transparent student model to mimic a black-box model, we intentionally include all features---even protected features like race and gender---specifically because we are interested in seeing what the models {\em could} learn from them. If, when the mimic model mimics the black-box model, it does not see any signal on the race or gender features and learns to model them as flat (zero) functions, this indicates whether the teacher model (the black-box model) did or did not use these features, but also if the teacher model exhibits race or gender bias even if the model did not use race or gender as inputs.

\section{Conclusion}
\label{sec:conclusion}
Our Distill-and-Compare approach to auditing black-box models was motivated by a realistic setting where access to the black-box model API is not available; only a data set labeled with the risk score as produced by the risk scoring model and the ground-truth outcome is available. The efficacy of the Distill-and-Compare audit approach increases when a model class that can be highly faithful to the black-box model and highly accurate at predicting the ground-truth outcomes is used, and when the audit data is not missing key features used in the black-box model. 

A key advantage of using transparent models to audit black-box models 
is that we do not need to know in advance what to look for. Many current auditing approaches focus on one or two protected features selected in advance, and thus are less likely to detect biases that are not \emph{a priori} known. The Distill-and-Compare audit approach, using transparent models, can hence be most useful for complicated, real-world data with unknown sources of biases. 

\newpage

\bibliographystyle{ACM-Reference-Format}
\bibliography{main}

\appendix
\section{A new confidence interval estimate for iGAM}
\label{sec:variance}
It is not trivial to estimate confidence intervals for nonparametric learners such as trees \cite{mentch2016quantifying}; iGAM's base learners are short trees. We employ a {\em bootstrap-of-little-bags} approach originally developed for bagged models in \cite{sexton2009standard} to estimate the variance of feature $x_i$'s contribution to the model, $h_i(x_i)$, and difference in feature $x_i$'s contribution to the mimic and outcome models, $sh_i(x_i) - oh_i(x_i)$.

Bootstrap-of-little-bags exploits two-level structured \\ cross-validation (e.g. 15\% of data points are selected for the test set, with the remaining 85\% split into training (70\%) and validation (15\%) sets). Repeating this inner splitting $L$ times and outer splitting $K$ times gives a total of $KL$ bags on which we train the mimic model and outcome model. Let $h^{lk}_i (x_i)$ be $x_i$'s feature contribution estimated by the model in the $l$th inner and $k$th outer fold. Its variance can then be estimated as \begin{equation*}\widehat{\mbox{Var}}(h_i(x_i)) = \frac{1}{K} \sum_{k=1}^K \left( \frac{1}{L} \sum_{l=1}^L h^{kl}_i(x_i) - \frac{1}{KL} \sum_{l=1}^l \sum_{k=1}^K h^{kl}_i(x_i) \right)^2\end{equation*}
and its mean $\overline{h_i(x_i)}$ is the average of the $KL$ models.

We can now construct pointwise confidence intervals for feature contributions to iGAM models. 
The confidence interval for feature $x_i$'s contribution to the model, $h_i(x_i)$, is $\overline{h_i(x_i)} \pm 1.96 \sqrt{ \widehat{\mbox{Var}}(h_i(x_i))}$
and the confidence interval for the difference in feature $x_i$'s contribution to the mimic and outcome models, $sh_i(x_i)-oh_i(x_i)$, is $\overline{sh_i(x_i) - oh_i(x_i)} \pm 1.96
\sqrt{\widehat{\mbox{Var}}(sh_i(x_i)) + \widehat{\mbox{Var}}(oh_i(x_i))-  2\widehat{\mbox{Cov}}(sh_i(x_i),oh_i(x_i))}$
with $\widehat{\mbox{Cov}}(sh_i(x_i),oh_i(x_i))$ also estimated using bootstrap-of-little-bags. 

This variance estimate is conservative (meaning it overestimates true variability), however, given that we are trying to detect differences, overestimating means we are less likely to mistake random noise for real differences. For large $K$ and $L$, consistency of this estimate was established in \cite{wager2017grf}. 

Note that are pointwise, not uniform, confidence intervals. That is, they capture the variability of the effect of age at age=50, not the entire effect of age. Uniform confidence intervals can be constructed by adjusting the critical value of the confidence interval. 



\begin{figure*}[htb]
\section{Calibration Plots}
\vspace{0.5cm}
\begin{center}
\begin{tabular}{ccccc}
\begin{turn}{90}\hspace{0.5cm}  \scriptsize{Fraction of Positive Outcomes, $p$}\end{turn}
\includegraphics[width=0.23\textwidth,height=1.4in]{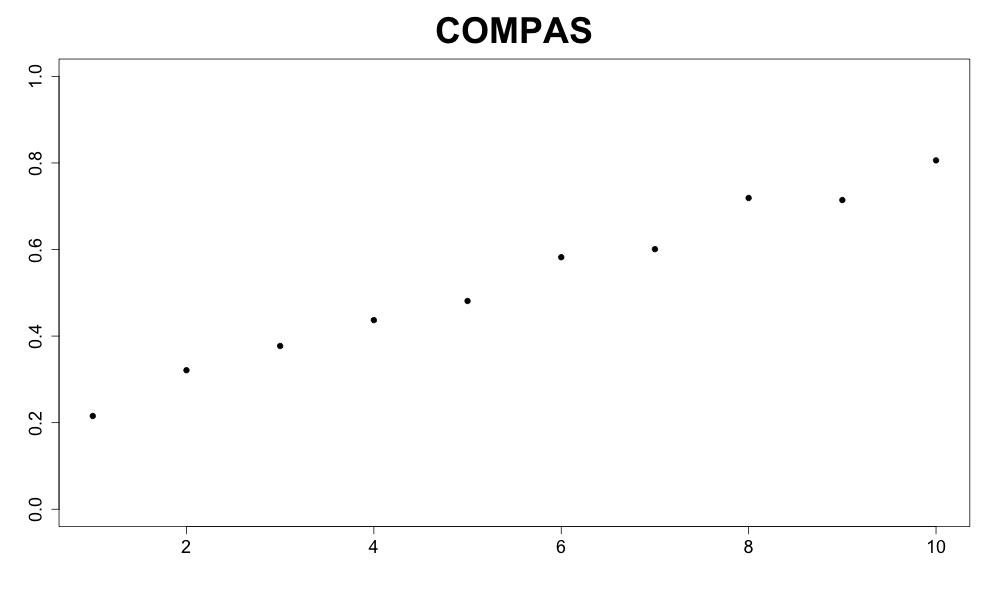} & \hspace*{-0.05in}
\includegraphics[width=0.23\textwidth,height=1.4in]{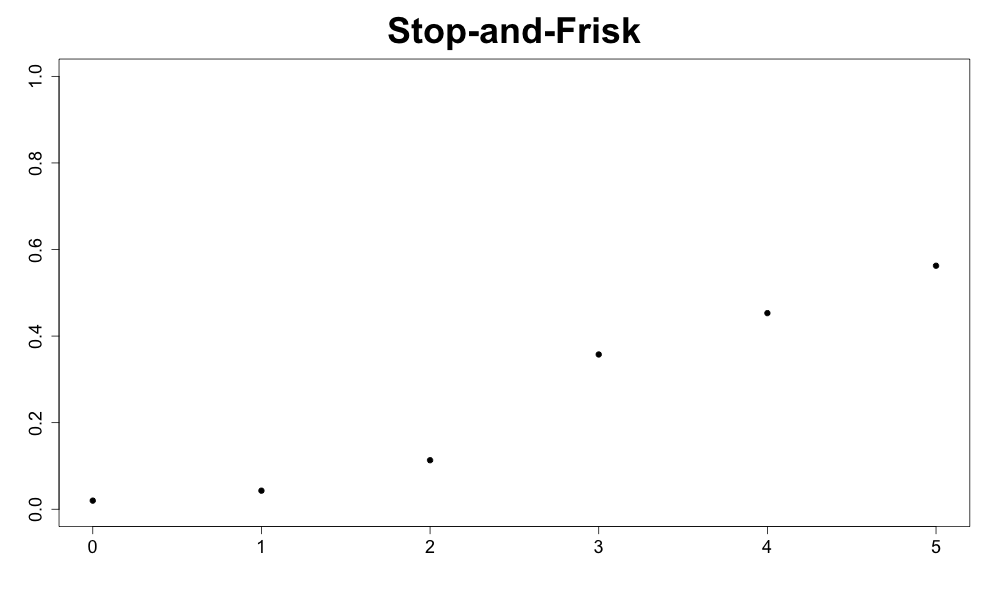} & \hspace*{-0.05in}
\includegraphics[width=0.23\textwidth,height=1.4in]{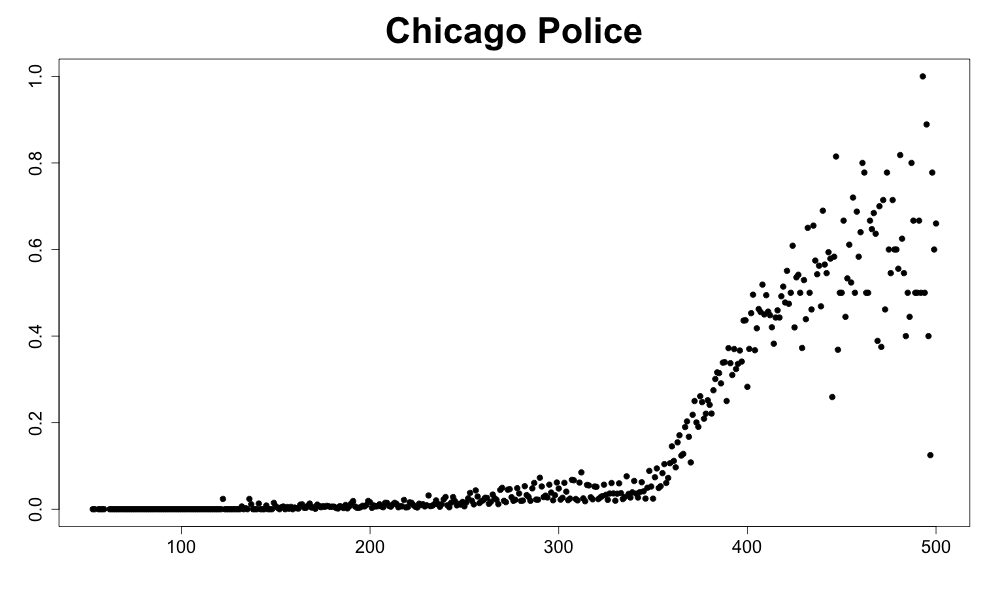} & \hspace*{-0.05in}
\includegraphics[width=0.23\textwidth,height=1.4in]{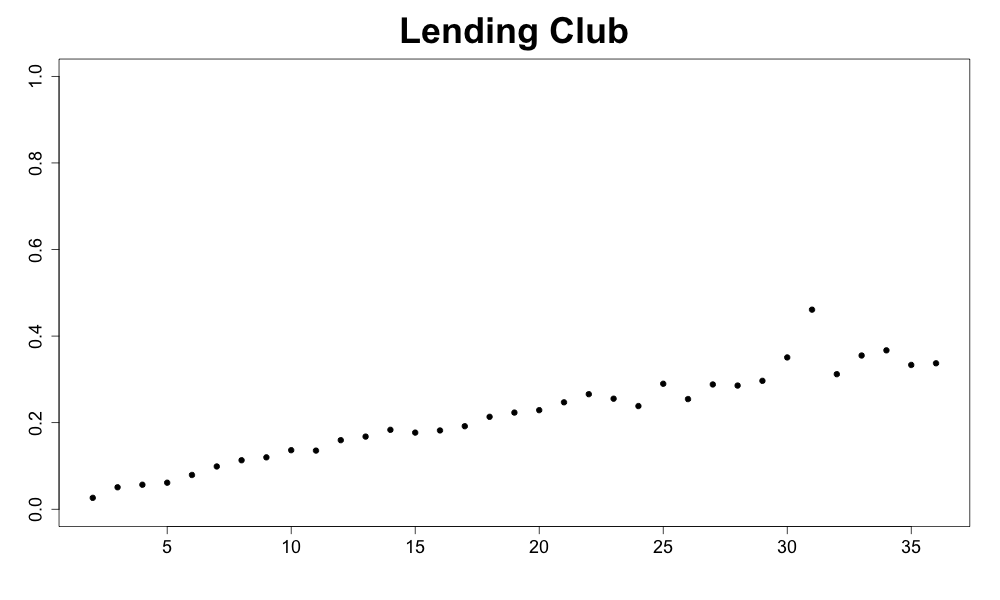} \hspace*{-0.05in} \\[-0.1in]
\begin{turn}{90}\hspace{0.5cm} \quad \quad \quad \scriptsize{logit($p$)}\end{turn}
\includegraphics[width=0.23\textwidth,height=1.4in]{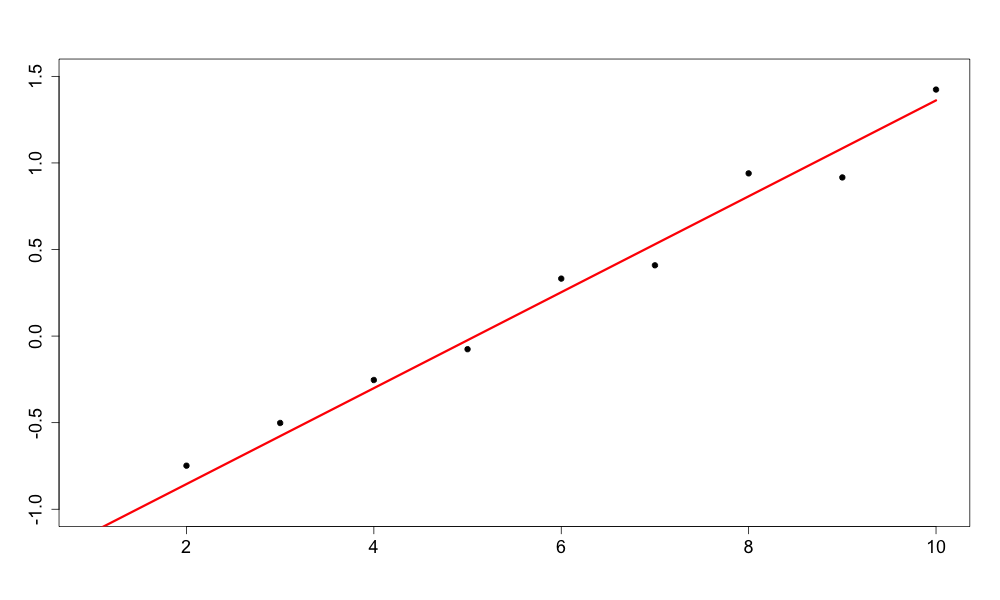} & \hspace*{-0.05in}
\includegraphics[width=0.23\textwidth,height=1.4in]{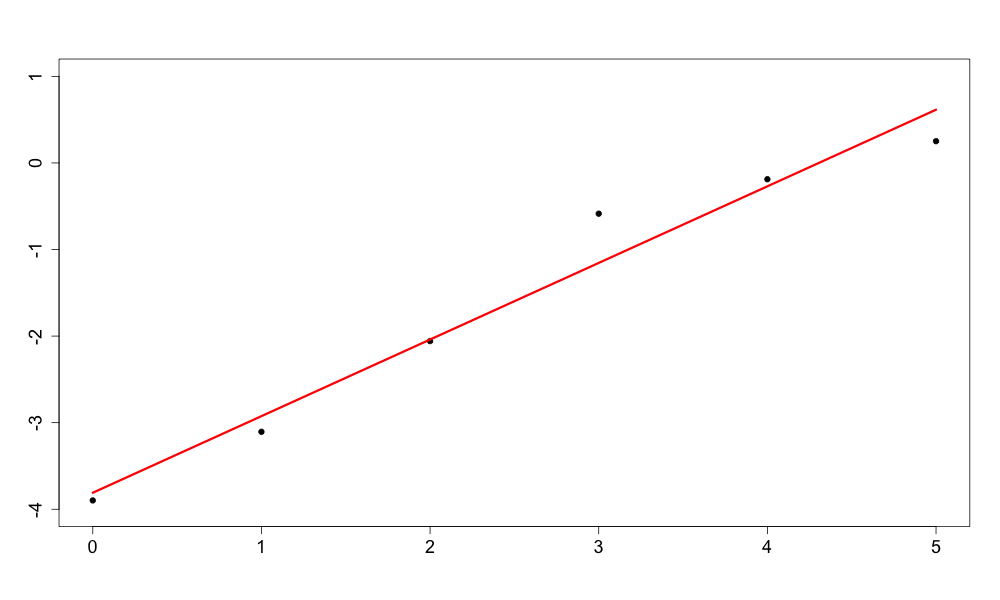} & \hspace*{-0.05in}
\includegraphics[width=0.23\textwidth,height=1.4in]{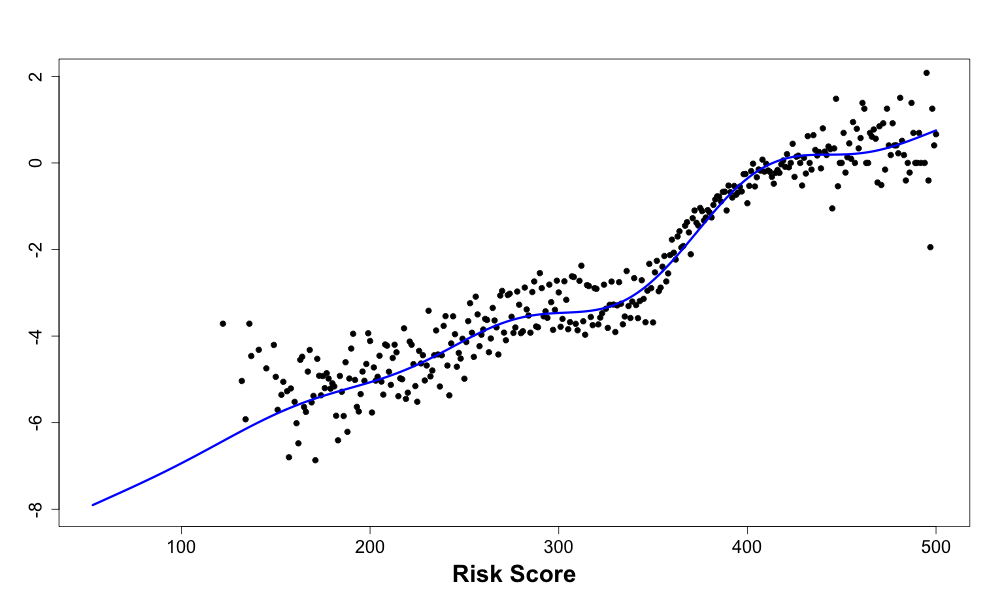} & \hspace*{-0.05in} 
\includegraphics[width=0.23\textwidth,height=1.4in]{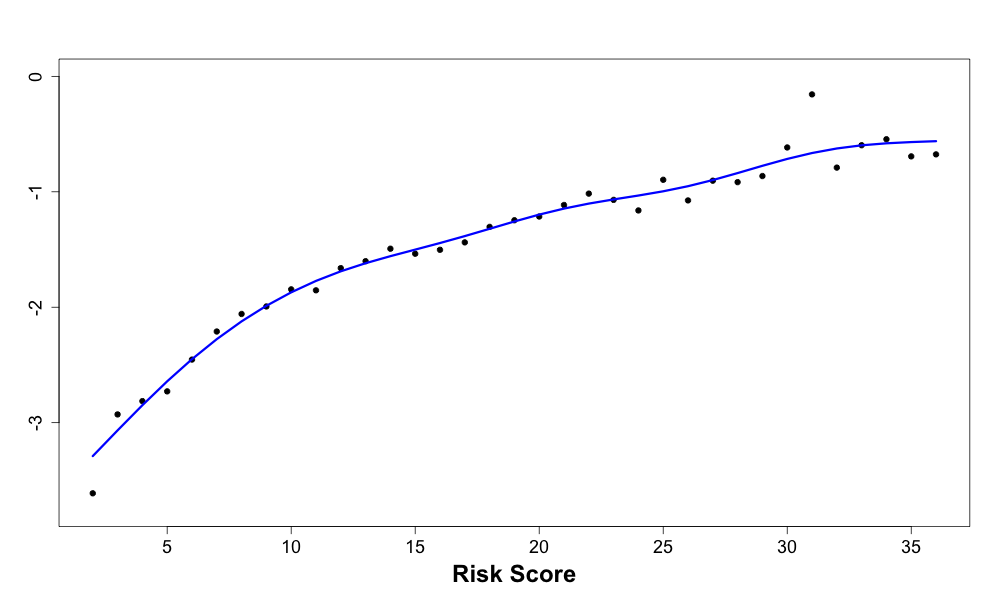} \hspace*{-0.05in} \\[-0.14in]
\begin{turn}{90}\hspace{0.5cm}\quad \scriptsize{\% Data}\end{turn}
\includegraphics[width=0.23\textwidth,height=0.85in]{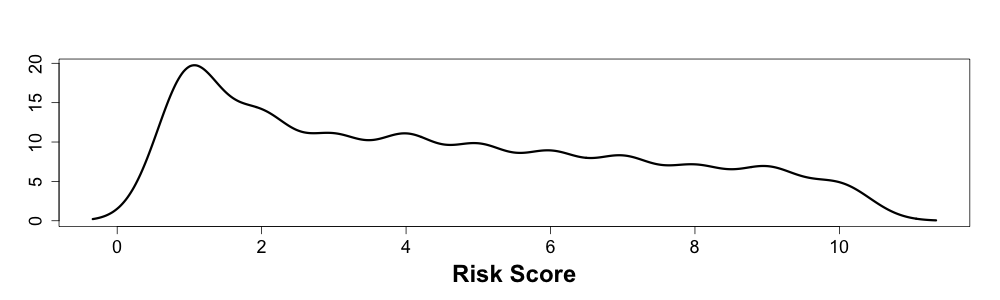} & \hspace*{-0.05in}
\includegraphics[width=0.23\textwidth,height=0.85in]{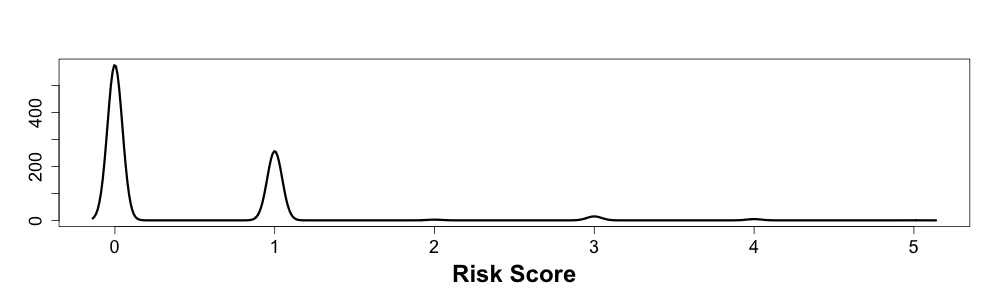} & \hspace*{-0.05in}
\includegraphics[width=0.23\textwidth,height=0.85in]{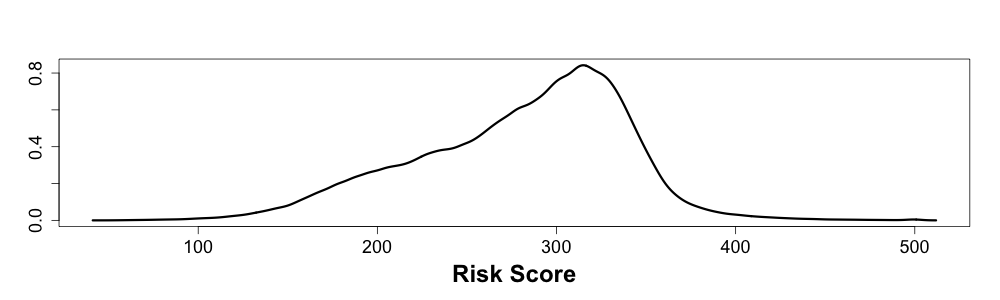} & \hspace*{-0.05in} 
\includegraphics[width=0.23\textwidth,height=0.85in]{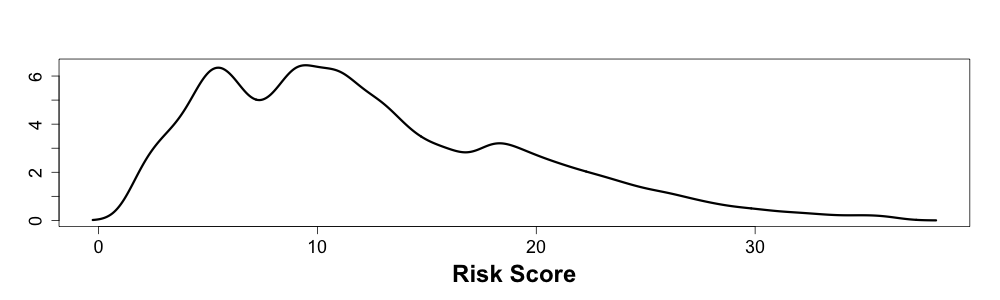} \hspace*{-0.05in} 
\end{tabular}
\end{center}
\vspace*{-0.05in}
\caption{Empirical probability (y-axis) vs. risk score (x-axis) for four risk scores on probability scale (top row) and logit probability scale (middle row). The red lines are best-fit straight lines. A good fit (COMPAS and Stop-and-Frisk) suggests that the risk score and outcomes logit probability (middle row) have a linear relationship; the mimic model can then be trained directly on the raw risk score. When the relationship is not linear (Chicago Police and Lending Club), the risk score must be calibrated (Section \ref{sec:calibration}). The blue monotonic curves are the learned nonlinear transformations. See Figure~\ref{fig:reliabilitycalibrated} for the transformed risk score. The bottom row describes the distribution of the risk scores.}
\label{fig:reliability}
\end{figure*}

\begin{figure*}[htb]
\vspace{1cm}
\begin{center}
\begin{tabular}{ccc}
\begin{turn}{90}\hspace{0.5cm}\quad \quad \quad \scriptsize{logit($p$)}\end{turn}
\includegraphics[width=0.23\textwidth,height=1.4in]{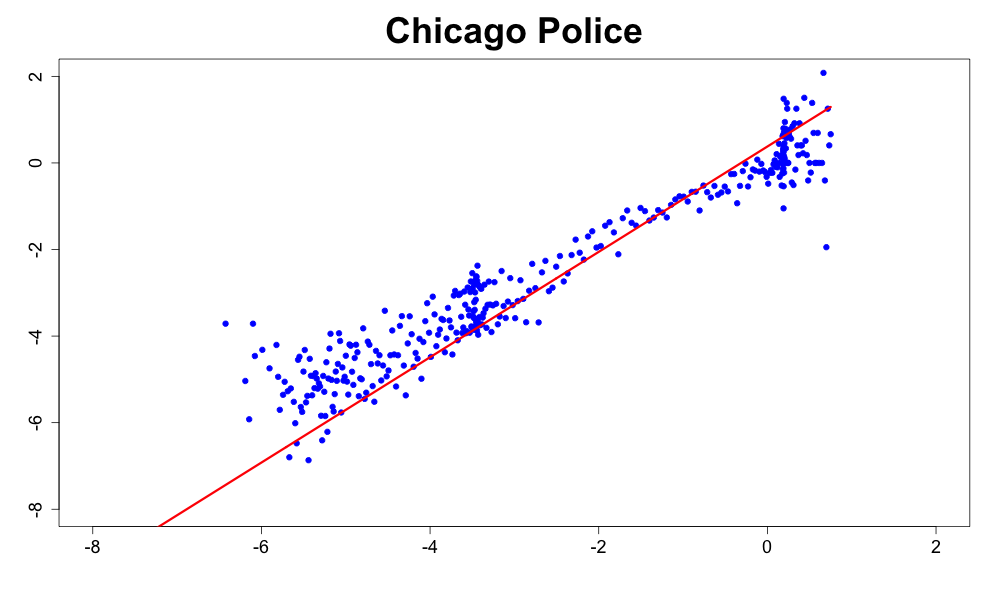} & \hspace*{-0.05in}
\includegraphics[width=0.23\textwidth,height=1.4in]{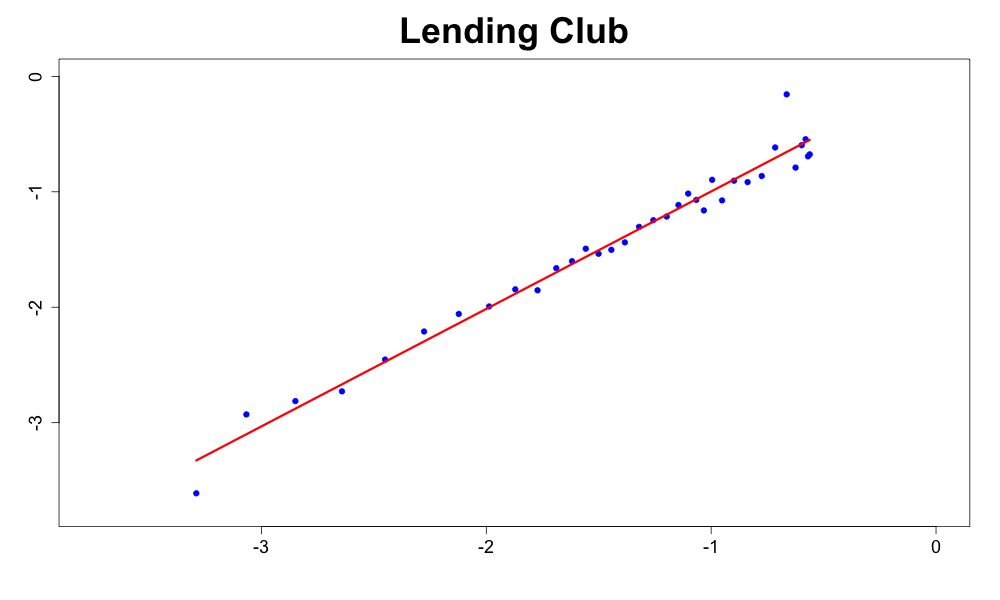} \hspace*{-0.05in} 
\end{tabular}
\end{center}
\vspace*{-0.05in}
\caption{Logit \emph{empirical} probability (y-axis) vs. risk score after applying nonlinear transformation (x-axis). The red lines are best-fit straight lines. A good fit suggests that the \emph{transformed} risk score and outcomes logit probability now have a linear relationship; the mimic model can now be traineed on the \emph{transformed} risk score. See Section \ref{sec:calibration} for details.
}
\label{fig:reliabilitycalibrated}
\end{figure*}

\end{document}